\documentclass[journal]{IEEEtran}

\usepackage{cite}

\usepackage{mahnigCEC}
\usepackage{microtype}
\usepackage{graphicx}
\usepackage{color,graphicx}
\usepackage{epstopdf} 
\usepackage{subfig}
\usepackage[hidelinks]{hyperref}
\usepackage{multirow}
\usepackage{array}
\usepackage[latin1]{inputenc}
\usepackage[T1]{fontenc}
\usepackage{bigstrut}
\usepackage{url}
\usepackage{lipsum}
\usepackage{graphicx}

\begin{document}

\title{MOEA/D-GM: Using probabilistic graphical models in MOEA/D for solving combinatorial optimization problems}

\author{Authors\thanks{Institution}}

\author{\IEEEauthorblockN{Murilo Zangari de Souza$^{1}$, Roberto Santana$^{2}$, Aurora Trinidad Ramirez Pozo$^{1}$, Alexander Mendiburu$^{2}$}

\IEEEauthorblockA{$^{1}$DInf - Federal University of Paran\'a,
CP: 19081, CEP 19031-970, Curitiba, Brazil\\ $^{2}$Intelligent Systems Group, University of the Basque Country UPV/EHU, San Sebasti\'an, Spain \\ Email: \{murilo.zangari, auroratrinidad\}@gmail.com, \{roberto.santana, alexander.mendiburu\}@ehu.eus}
}


\maketitle

\begin{abstract}
 Evolutionary algorithms based on modeling the statistical dependencies (interactions) between the variables have been proposed to solve a wide range of complex problems. These algorithms learn and sample probabilistic graphical models able to encode and exploit the regularities of the problem. This paper investigates the effect of using probabilistic modeling techniques as a way to enhance the behavior of MOEA/D framework. MOEA/D is a decomposition based evolutionary algorithm that decomposes an multi-objective optimization problem (MOP) in a number of scalar single-objective subproblems and optimizes them in a collaborative manner. MOEA/D framework has been widely used to solve several MOPs. The proposed algorithm, MOEA/D using probabilistic Graphical Models (MOEA/D-GM) is able to instantiate both univariate and multi-variate probabilistic models for each subproblem. To validate the introduced framework algorithm, an experimental study is conducted on a multi-objective version of the deceptive function \textit{Trap5}. The results show that the variant of the framework (MOEA/D-Tree), where tree models are learned from the matrices of the mutual information between the variables, is able to capture the structure of the problem. MOEA/D-Tree is able to achieve significantly better results than both MOEA/D using genetic operators and MOEA/D using univariate probability models in terms of approximation to the \textit{true} Pareto front. 
\end{abstract}


\IEEEpeerreviewmaketitle

\section{Introduction}


Several real-world problems can be stated as multi-objective optimization problems (MOPs) which have two or more objectives to be optimized. Very often these objectives conflict with each other. Therefore, no single solution can optimize these objectives at the same time. Pareto optimal solutions are of very practical interest to decision makers for selecting a final preferred solution. Most MOPs may have many or even infinite optimal solutions and it is very time consuming (if not impossible) to obtain the complete set of optimal solutions \cite{Coello:2007}. 

Since early nineties, much effort has been devoted to develop evolutionary algorithms for solving MOPs. Multi-objective evolutionary algorithms (MOEAs) aim at finding a set of representative Pareto optimal solutions in a single run \cite{Coello:2007,Zhang_and_Li:2007,Raplacement:2014}. 
 
Different strategies have been used as criteria to maintain a population of optimal non-dominated solutions, or Pareto set (\textit{PS}), and consequently finding an approximated Pareto front (\textit{PF}). Popular strategies include:\footnote{There exist algorithms, such as NSGAIII \cite{Deb:2014b}, that combine the ideas from both decomposition based and Pareto dominance based.} (i) Pareto dominance based; (ii) Indicator based and (iii) Decomposition based (also called, scalarization function based) \cite{Raplacement:2014,MOEADforMany:2014}. 

Zhang and Li \cite{Zhang_and_Li:2007} proposed a decomposition based algorithm called MOEA/D (multi-objective Evolutionary Algorithm Based on Decomposition) framework, which decomposes a MOP into a number of single-objective scalar optimization subproblems and optimizes them simultaneously in a collaborative manner using the concept of neighborhood between the subproblems.

Current research on MOEA/D are various and include the extension of these algorithms to continuous MOPs with complicated Pareto sets \cite{ComplicatedPareto:2008}, many-objective optimization problems \cite{Ishibushi:2013,Tan_et_al:2013,MOEADforMany:2014}, methods to parallelize the algorithm \cite{Nebro_and_Durillo:2010}, incorporation of preferences to the search \cite{Pilat_and_Neruda:2015}, automatic adaptation of weight vectors \cite{Qi_et_al:2014}, new strategies of selection and replacement to balance convergence and diversity \cite{Raplacement:2014}, hybridization with local searches procedures \cite{moeadtabu:2014, Ke_and_Zhang:2014}, etc.

However, it has been shown that traditional EA operators fail to properly solve some problems when certain characteristics are present in the problem such as deception \cite{Larranaga_et_al:2012}. A main reason for this shortcoming is that, these algorithms do not consider the dependencies between the variables of the problem \cite{Larranaga_et_al:2012}. To address this issue, evolutionary algorithms that (instead of using classical genetic operators) incorporate machine learning methods have been proposed. These algorithms are usually referred to as Estimation of Distribution Algorithms (EDAs). In EDAs, the collected information is represented using a probabilistic model which is later employed to generate (sample) new solutions \cite{Larranaga_et_al:2012,Lozano_et_al:2005,Muhlenbein_and_Paas:1996}. 

EDAs based on modeling the statistical dependencies between the variables of the problem have been proposed as a way to encode and exploit the regularities of complex problems. Those EDAs use more expressive probabilistic models that, in general, are called probabilistic graphical models (PGMs) \cite{Larranaga_et_al:2012}. A PGM comprises a graphical component representing the conditional dependencies between the variables, and a set of parameters, usually tables of marginal and conditional probabilities. Additionally, the analysis of the graphical components of the PGMs learned during the search can provide information about the problem structure. MO-EDAs using PGMs have been also applied for solving different MOPs  \cite{Pelikan_et_al:2006a,Marti_et_al:2010,Karshenas_et_al:2014}.

In the specified literature of MO-EDAs, a number of algorithms that integrate to different extents the idea of probabilistic modeling into MOEA/D have been previously proposed \cite{Shim:2012,Zhou:2013,Wang:2015,Giagkiozis_et_al:2012,Zapotecas_et_al:2015}. Usually, these algorithms learn and sample a probabilistic model for each subproblem. Most of them use only univariate models, which are not able to represent dependencies between the variables. 

 In this paper, we investigate the effect of using probabilistic modeling techniques as a way to enhance the behavior of MOEA/D. We propose a MOEA/D framework able to instantiate different PGMs. The general framework is called MOEA/D Graphical Models (MOEA/D-GM). The goals of this paper are: (i) Introduce MOEA/D-GM as a class of MOEA/D algorithms that learn and sample probabilistic graphical models defined on discrete domain; (ii) To empirically show that MOEA/D-GM can improve the results of traditional MOEA/D for deceptive functions; (iii) To investigate the influence of modeling variables dependencies in different metrics used to estimate the quality of the Pareto front approximations and (iv) to show, for the first time, evidence of how the problem structure is captured in the models learned by MOEA/D-GM.
 



The paper is organized as follows. In the next section some relevant concepts used in the paper are introduced. Sections~\ref{sec:MOEA/D} and~\ref{sec:EDAs} respectively explain the basis of  MOEA/D and the EDAs used in the paper.  Section~\ref{sec:MOEADGM} introduces the MOEA/D-GM and explains the enhancements required by EDAs in order to efficiently work within the MOEA/D context. In Section~\ref{sec:DECEPTION} we discuss the class of functions which are the focus of this paper, multi-objective deceptive functions. We explain the rationale of applying probabilistic modeling for these functions. Related work is discussed in detail in Section~\ref{sec:RELWORK}. The experiments that empirically show the behavior of MOEA/D-GM are described in Section~\ref{sec:EXPE}. The conclusions of our paper and some possible trends for future work are presented in Section~\ref{sec:CONCLU}. 

\section{Preliminaries}  

 Let ${\bf{X}}=(X_1,\ldots ,X_n)$ denote a vector of discrete random variables. ${\mathbf{x}}=(x_1,\ldots ,x_n)$ is used to denote an assignment to the variables and $\mathbf{x} \in \{0,1\}^{n}$. A population is represented as a set of vectors $x^1,\dots, x^N$ where $N$ is the size of the population. Similarly, $x^i_j$ represents the assignment to the $j$th variable of the $i$th solution in the population. 

A general MOP can be defined as follows \cite{Coello:2007}:
\begin{eqnarray}
\label{mop}
\textit{minimize (or maximize)		} F(\mathbf{x}) = (f_{1}(\mathbf{x}),...,f_{m}(\mathbf{x}))
 \\ \nonumber subject \ to \ \mathbf{x} \in \Omega
\end{eqnarray}
where $\mathbf{x} = (x_{1}, x_{2},\dots x_{n})^{T}$ is the decision variable vector of size $n$, $\Omega$ is the \textit{decision space}, $F : \Omega \rightarrow R^{m}$ consists of \textit{m} objective functions and $R^{m}$ is the \textit{objective space}.

Pareto optimality is used to define the set of optimal solutions.

Pareto optimality\footnote{This definition of domination is for minimization. For maximization, the inequalities should be reversed.} \cite{Coello:2007}: Let $\mathbf{x}, \mathbf{y} \in \Omega$, $\mathbf{x}$ is said to \textit{dominate} $\mathbf{y}$ if and only if $f_{l}(\mathbf{x}) \leq f_{l}(\mathbf{y})$ for all $l \in \{1,..,m\}$ and $f_{l}(\mathbf{x}) < f_{l}(\mathbf{y})$ for at least one $l$. A solution $\mathbf{x}^{*} \in \Omega$ is called \textit{Pareto optimal} if there is no other $\mathbf{x} \in \Omega$ which \textit{dominates} $\mathbf{x}^{*}$. The set of all the Pareto optimal solutions is called the \textit{Pareto set (PS)} and the solutions mapped in the \textit{objective space} are called \textit{Pareto front (PF)}, i.e., $PF = \{F(\mathbf{x}) | \mathbf{x} \in PS \}$. In many real life applications, the \textit{PF} is of great interest to decision makers for understanding the trade-off nature of the different objectives and selecting their preferred final solution.

\section{MOEA/D} \label{sec:MOEA/D}  

MOEA/D \cite{Zhang_and_Li:2007} decomposes a MOP into a number $(N)$ of scalar single objective optimization subproblems and optimizes them simultaneously in a collaborative manner using the concept of neighborhood between the subproblems. Each subproblem $i$ is associated to a weight vector $\lambda^{i} = (\lambda^{i}_{1},...,\lambda^{i}_{m})^{T}$ and the set of all weight vectors is $\{\lambda^{1},...,\lambda^{N}\}$. 

Several decomposition approaches have been used in the MOEA/D \cite{Decomposition:2013}. The \textit{Weighted Sum} and \textit{Tchebycheff} are two of the most common approaches used, mainly in combinatorial problems \cite{Decomposition:2013,Zhang_and_Li:2007}.  

 Let $\lambda = (\lambda_{1},...,\lambda_{m})^{T}$ be a weight vector, where $\sum_{l=1}^{m} \lambda_{l} = 1$ and $\lambda_{l} \geq 0$ for all $l=1,...,m$. 
 
\textit{Weighted Sum Approach:} The optimal solution to the following scalar single-optimization problem is defined as:
\begin{eqnarray}
\label{decompo}
\textit{minimize (or maximize)	} \large  g^{ws}(\mathbf{x}|\lambda) =\sum_{l=1}^{m} \lambda_{l}f_{l}(\mathbf{x}) \nonumber 
\\
\textit{subject to		}  \mathbf{x} \in \Omega
\end{eqnarray}
where we use $g^{ws}(\mathbf{x}|\lambda)$ to emphasize that $\lambda$ is a weight vector in this objective function. $\mathbf{x}$ is a Pareto optimal solution to (\ref{mop}) if the \textit{PF} of (\ref{mop}) is convex (or concave).

\textit{Tchebycheff approach:} The optimal solution to the following scalar single-optimization problems is defined as:
 \begin{eqnarray}\label{TB}
minimize \ \ g^{te}(\mathbf{x}|\lambda ,z^{*}) = \max_{1 \leq l \leq m} \{\lambda_{l} | f_{l}(\mathbf{x}) - z^{*}_{l} | \}; \\  
 subject \ to \ \mathbf{x} \in \Omega \nonumber
\end{eqnarray}
where $z^{*} = (z^{*}_{1}, ..., z^{*}_{m})^{T}$ is the reference point, i.e., $z^{*}_{l} = \max\{f_{l}(\mathbf{x})|\mathbf{x} \in \Omega\}$ for each $l = 1, ...,m$. For each Pareto optimal solution $x^{*}$ there exists a weight vector $\lambda$ such that $x^{*}$ is the optimal solution of (\ref{TB}) and each optimal solution of (\ref{TB}) is Pareto optimal of (\ref{mop}).

The neighborhood relation among the subproblems is defined in MOEA/D. The neighborhood for each subproblem $i$, $B(i) = \{i_{1},...,i_{T}\}$,  is defined according to the Euclidean distance between its weight vector and the other weight vectors. 
The relationship of the neighbor subproblems is used for the selection of parent solutions and the replacement of old solutions (also called update mechanism). The size of the neighborhood used for selection and replacement plays a vital role in MOEA/D to exchange information among the subproblems \cite{Raplacement:2014,MOEADforMany:2014}. Moreover, optionally, an external population \textit{EP} is used to maintain all Pareto optimal solutions found during the search. Algorithm~\ref{alg:moead} (adapted from \cite{Raplacement:2014}) presents the pseudo-code of the general MOEA/D which serves as a basis for this paper.

\begin{BAlgo}{General MOEA/D framework}
 \label{alg:moead} \small
 \item  $[Pop, T_{s}, T_{r}, EP]$ $\leftarrow$ Initialization() $//$ $Pop$ is the initial population of size $N$. Set $B(i)=T_{s}$ and $R(i)=T_{r}$, where $B(i)$ and $R(i)$ are, respectively, the  neighborhood size for selection and replacement. $EP$ is the external population.  
  \item {While a termination condition not met}  
 \item \T {For each subproblem $i \in 1,...,N$ at each generation}
  \item  \TT {$\bf{y} \leftarrow$ Variation($B(i)$)} 
  \item \TT {Evaluate $F(\bf{y})$ using the fitness function}
 \item  \TT {Update\_Population($Pop, \bf{y},$ $R(i)$)}
\item \TT {Update\_$EP$($F(\bf{y}$),\textit{EP})} 
\item Return $Pop, EP$
\end{BAlgo}

\subsubsection{Initialization} The $N$ weight vectors $\lambda^{1},..,\lambda^{N}$ are set. The Euclidean distance between any two weight vectors is computed. For each subproblem $i$, the set of neighbors for the selection step $B(i)$ and the update step $R(i)$ are initialized with the $T_{s}$ and $T_{r}$ closest neighbors respectively according to the Euclidean distance. The initial population $ Pop = x^{1},...,x^{N}$ is generated in a random way and their corresponding fitness function $F(x^{1}),...,F(x^{N})$ are computed. The external Pareto $EP$ is initialized with the non-dominated solutions from the initial population $Pop$.

\subsubsection{Variation}  The reproduction is performed using $B(i) = \{i_{1},...,i_{T_{s}} \}$ to generate a new solution $\mathbf{y}$. The conventional algorithm selects two parent solutions from $B(i)$ and applies \textit{crossover} and \textit{mutation} to generate $\bf{y}$.

\subsubsection{Update Population} This step decides which subproblems should be updated from $R(i)=\{i_{1},...,i_{T_{r}}\}$. The current solutions of these subproblems are replaced by $\mathbf{y}$ if $\mathbf{y}$ has a better aggregation function value $g^{te}(\mathbf{y}|\lambda^{r})$ than $g^{te}(x^{r}|\lambda^{r})$, $x^{r} \in R(i)$

\subsubsection{Update EP} This step removes from $EP$ the solutions dominated by $\mathbf{y}$ and adds $\mathbf{y}$ if no solution dominates it.

\subsection{Improvements on MOEA/D framework}
 
Wang et. al \cite{Raplacement:2014} have reported that different problems need different trade-offs between diversity and convergence, which can be controlled by the different mechanisms and parameters of the algorithm. As so far, most proposed MOEA/D versions adopted the $(\mu + 1)$-selection/variation scheme, which selects $\mu$ individuals from a population and generates 1 offspring \cite{Raplacement:2014}. Different strategies for selection and replacement have been proposed \cite{Raplacement:2014,MOEADforMany:2014}. In this paper, the replacement proposed by \cite{ComplicatedPareto:2008} is used. In this strategy, the maximal number of solutions replaced by a new solution $\mathbf{y}$ is bounded by $n_{r}$, which should be set to be much smaller than $T_{r}$. This replacement mechanism prevents one solution having many copies in the population. 

\section{Estimation of distribution algorithms} \label{sec:EDAs}

EDAs \cite{Larranaga_et_al:2012,Lozano_et_al:2005,Muhlenbein_and_Paas:1996} are stochastic population-based optimization algorithms that explore the space of candidate solutions by sampling a probabilistic model constructed from a set of selected solutions found so far. Usually, in EDAs, the population is ranked according to the fitness function. From this ranked population, a subset of the most promising solutions are selected by a \textit{selection operator} (such as: \textit{Truncation selection} with truncation threshold $t=50\%$). The algorithm then constructs a probabilistic model which attempts to estimate the probability distribution of the selected solutions. Then, according to the probabilistic model, new solutions are sampled and incorporated into the population, which can be entirely replaced \cite{Pelikan:2011}. The algorithm stops when a termination condition is met such as the number of generations.

We work with positive distributions denoted by $p$. $p(x_I)$ denotes  the  marginal probability for ${\bf{X}}_I={\bf{x}}_I$.   $p(x_j \mid x_k)$ denotes the conditional probability distribution of $X_j=x_j$ given $X_k=x_k$. The set of selected promising solutions is represented as $S$. Algorithm~\ref{alg:EDA} presents the general EDA procedure.

\begin{BAlgo}{General EDA} 
 \label{alg:EDA} \small
 \item { $Pop \leftarrow $ Generate $N$ solutions randomly}
 \item While a termination condition not met
 \item \T {For each solution $x^{i}$ compute its fitness function $F(x^{i})$} 
 \item \T { Select the most promising solutions $S$ from $Pop$}
 \item \T {Build a probabilistic model $\mathcal{M}$ of solutions in $S$}
\item \T {Generate new candidate solutions sampling from $\mathcal{M}$ and add them to $Pop$} 		
 \end{BAlgo}

The way in which the learning and sampling components of the algorithm are implemented are also critical for their performance and computational cost \cite{Larranaga_et_al:2012}. In the next section, the univariate and \textit{tree-based} probabilistic models are introduced.  

\subsection{Univariate probabilistic models}

In the univariate marginal distribution (or univariate probabilistic model), the variables are considered to be independent, and the probability of a solution is the product of the univariate probabilities for all the variables:

\begin{equation}
 p_{u}({\bf{x}}) =   \prod_{j=1}^{n}  p(x_{j}) \label{eq:UNIV}
\end{equation}

One of the simplest EDAs that use the univariate model is the univariate marginal distribution algorithm (UMDA) \cite{Muhlenbein:1997}. UMDA uses a probability vector $p_{u}(\mathbf{x})$ as the probabilistic model, where $p(x_{j})$ denotes the univariate probability associated to the corresponding discrete value. In this paper we focus on binary problems.  To learn the probability vector for these problems, each $p(x_{j})$ is set to the proportion of "1s" in the selected population $S$. To generate new solutions, each variable is independently sampled. Random values are assigned to the variables and following the corresponding univariate distribution.

The population-based incremental learning (PBIL) \cite{Baluja:1994}, like UMDA, uses the probabilistic model in the form of a probability vector $p_{u}(\mathbf{x}$). The initial probability of a "1" in each position $p(x_{j})$ is set to $0.5$. The probability vector is updated using each  selected solution $\mathbf{x}$ in $S$. For each variable, the corresponding entry in the probability vector is updated by:
\begin{equation}
p(x_{j}) = (1.0 - \alpha) p(x_{j}) + (\alpha * x_{j})
\end{equation}
where $\alpha$ is the learning rate specified by the user.

To prevent premature convergence, each position of the probability vector is slightly varied at each generation based on a mutation rate parameter \cite{Baluja:1994,Pelikan:2011}. Recently, it has been acknowledged \cite{Zangari_et_al:2015} that implementations of the probabilistic vector update mechanism in PBIL are in fact different, and produce an important variability in the behavior of PBIL.  

Univariate approximations are expected to work well for functions that can be additively decomposed into  functions of order one (e.g. $g({\bf{x}})= \sum_j x_j$). Also, other non additively decomposable functions can be solved with EDAs that use univariate models (e.g. $g({\bf{x}})=   \prod_j x_j + \sum_j x_j $) \cite{Muhlenbein_et_al:1999}.

\subsection{Tree-based models}
	
	Different from the EDAs that use univariate models, some EDAs can assume dependencies between the decision variables. In this case, the probability distribution is represented by a PGM. 
	
	\textit{Tree-based} models \cite{Pelikan_and_Muhlenbein:1999} are PGMs capable to capture some \textit{pairwise} interactions between variables. In a tree model, the conditional probability of a variable may only depend on at most one other variable, its parent in the tree. 
	
The probability distribution $p_{\mathcal{T}} ({\bf{x}})$ that is conformal with a tree model is defined as:
\begin{equation}
  p_{\mathcal{T}} ({\bf{x}}) =\prod_{j=1}^{n} p(x_j|pa(x_j)),
\end{equation}
 where $pa(X_j)$  is the parent of $X_j$ in the tree, and $p(x_j|pa(x_j))=p(x_j)$ when $pa(X_j)=\emptyset$, i.e. $X_j$ is a root of the tree.	
	
	The bivariate marginal distribution algorithm (BMDA) proposed in \cite{Pelikan_and_Muhlenbein:1999} uses a model based on a set of mutually independent trees (a forest)\footnote{For convenience of notation, this set of mutually independent trees is refereed as Tree.}. In each iteration of the algorithm, a tree model is created and sampled to generate new candidate solutions based on the conditional probabilities learned from the population.

The algorithm Tree-EDA proposed in \cite{Santana_et_al:2001b} combines features from algorithms introduced in \cite{Baluja_and_Davies:1997r} and~\cite{Pelikan_and_Muhlenbein:1999}. In Tree-EDA, the learning procedure works as follows:
\begin{itemize}
\item Step 1: Compute the univariate and bivariate marginal frequencies $p_{j}(x_{j})$ and $p_{jk}(x_{j}|x_{k})$ using the set of selected promising solutions $S$;
\item Step 2: Calculate the matrix of mutual information using the univariate and bivariate frequencies;
\item Step 3: Calculate the maximum weight spanning tree from the mutual information. Compute the parameters of the model.
\end{itemize} 

The idea is that by computing the maximum weight spanning tree from the matrix of mutual information the algorithm will be able to capture the most relevant bivariate dependencies between the problem variables.

Details on EDAs that use the tree models can be obtained from \cite{Baluja_and_Davies:1997r,Pelikan_and_Muhlenbein:1999}. More details  on the use of PGMs for probabilistic modeling in EDAs can be obtained from \cite{Larranaga_et_al:2012}.

\subsection{Optimal Mutation Rate for EDAs}

Most of EDAs do not use any kind of stochastic mutation. However, for certain problems, lack of diversity in the population is a critical issue that can cause EDAs to produce poor results due to premature convergence. As a remedy to this problem, in \cite{Prior:2001}, the authors propose the use of Bayesian \textit{priors} as an effective way to introduce a  mutation operator into UMDA. Bayesian priors are used for the computation of the univariate probabilities in such a way that the computed probabilities will include a mutation-like effect. In this paper we use the Bayesian prior as a natural way to introduce mutation in the MOEA/D-GM. In the following this strategy is described.  

There are two possibilities to estimate the probability of "head" of a biased coin. The maximum-likelihood estimate counts the number of occurrences of each case. With $m$ times "head" in $l$ throws, the probability is estimated as $p=\frac{m}{l}$. The Bayesian approach assumes that the probability of "head" of a biased coin is determined by an unknown parameter $\theta$. Starting from an \textit{a priori} distribution of this parameter. Using the Bayesian rule the univariate probability is computed as $p = \frac{m + r}{l + 2r}$, and the so called \textit{hyperparameter} $r$ has to be chosen in advance. 

To relate the Bayesian \textit{prior} to the mutation, the authors used the following theorem: \textit{For binary variables, the expectation value for the probability using a Bayesian \textit{prior} with parameter $r$ is the same as mutation with mutation rate} $\mu = \frac{r}{l + 2r}$ \textit{and using the maximum likelihood estimate} \cite{Prior:2001}.

\section{MOEA/D-GM: An MOEA/D algorithm with graphical models} \label{sec:MOEADGM}

 Algorithm~\ref{Alg:moead-gm} shows the steps of the proposed MOEA/D-GM framework. First, the initialization procedure generates $N$ initial solutions and the external Pareto (\textit{EP}) is initialized with the non-dominated solutions. Within the main while-loop, in case the termination criteria are not met, for each subproblem $i$, a probabilistic model is learned  using as base population all solutions in the neighborhood $B(i)$. Then, the sampling procedure is used to generate a new solution from the model. The new solutions is used to update the parent population according to an elite-preserving mechanism. Finally, the new solution is used to update \textit{EP} as described in Algorithm 1.
 
\begin{BAlgo}{MOEA/D-GM}
 \label{Alg:moead-gm} \scriptsize
\item  $[Pop, T_{s}, T_{r}, EP]$ $\leftarrow$ Initialization() $//$ $Pop$ is the initial population of size $N$ that is randomly generated. Set $B(i)=T_{s}$ and $R(i)=T_{r}$, where $B(i)$ and $R(i)$ are, respectively, the  neighborhood size for selection and replacement. $EP$ is the external population.  
 \item {While a termination condition not met}  
 \item \T{For each subproblem $i \in 1,...,N$ at each generation}
 \item \TT{\textbf{Learning()}}
 \item \TTT{Case GA: Choose two parent solutions $x^{1} x^{2}$ from $B(i)$}
 \item \TTT{Case UMDA: Learn a probabilistic vector $p^{i}$ using the solutions from $B(i)$} 
  \item \TTT{Case PBIL: Learn an incremental probabilistic vector $p^{i}$ using the solutions from $B(i)$}   
 \item \TTT{Case Tree-EDA: Learn a \textit{tree} model $p_{\mathcal{T}}^{i}$ using the solutions from $B(i)$}
 \item \TT{\textbf{Sampling()} $//$ Try up to $T_{s}$ times to sample a new solution $\bf{y}$ that is different from any solution from $B(i)$}
 \item \TTT{Case GA: Apply the \textit{crossover} and mutation to generate $\bf{y}$}
 \item \TTT{Case PBIL or UMDA: Sample $\bf{y}$ using the probability vector $p^{i}$}
 \item \TTT{Case Tree-EDA: Sample $\bf{y}$ using the \textit{tree} model $p_{\mathcal{T}}^{i}$}
 \item \TT{Compute the fitness function $F(\bf{y})$}
 \item \TT{Update\_Population($Pop, \bf{y},$ $R(i)$)}
 \item \TT{Update\_$EP$($F(\bf{y})$,\textit{EP})}
 \item Return $Pop, EP$
\end{BAlgo}

A simple way to use probabilistic models into MOEA/D is learning and sampling a probabilistic model for each scalar subproblem $i$ using the set of closest neighbors as the selected population. Therefore, at each generation, the MOEA/D-GM keeps $N$ probabilistic models. 

The EDAs presented in the previous section: (i) UMDA \cite{Muhlenbein:1997}, (ii) PBIL \cite{Baluja:1994} and (iii) Tree-EDA \cite{Santana_et_al:2001b} can be instantiated in the framework. Also, the genetic operators (\textit{crossover} and mutation), that are used in the standard MOEA/D, can be applied using the learning and sampling procedures.

The general MOEA/D-GM  allows the introduction of other types of probabilistic graphical models (PGMs) like Bayesian networks \cite{Pelikan_et_al:1999} and Markov networks \cite{Shakya_and_Santana:2012}. Different classes of PGMs can be also used for each scalar subproblem but we do not consider this particular scenario in the analysis presented here. 

   

 Moreover, the MOEA/D-GM introduces some particular features that are explained in the next section.
 
\subsubsection{Learning models from the complete neighborhood}

 Usually in EDAs, population sizes are large and the size of the selected population can be as high as the population size. In MOEA/D, each subproblem has a set of neighbors that plays a similar role to the selected population\footnote{Usually, the size of the neighborhood is $N \times 0.1$. Although, as in \cite{ComplicatedPareto:2008}, the propose MOEA/D has a low probability to select two parent solutions from the complete population.}. Therefore, the main difference between the MOEA/D-GM and the EDAs presented in section \ref{sec:EDAs} is that, in the former, instead of selecting a subset of individuals based on their fitness to keep a unique probabilistic model, a probabilistic model is computed over the neighborhood solutions of each scalar subproblem.
 




 \subsubsection{Diversity preserving sampling}

  In preliminary experiments we detected that one cause for early convergence of the algorithm was that solutions that were already in the population were newly sampled. Sampling solutions already present in the population is also detrimental in terms of efficiency since these solutions have to be evaluated. As a way to avoid this situation, we added a simple procedure in which each new sampled solution is tested for presence in the neighborhood $B(i)$ of each subproblem $i$.
 
If a new sampled solution $\mathbf{y}$ is equal to a parent solution from $B(i)$ then the algorithm discards the solution and samples a new one until a different solution has been found or a maximum number of trials is reached. This procedure is specially suitable to deal with expensive fitness functions. The maximum number of tries can be specified by the user. We call the sampling that incorporates this verification procedure as \emph{diversity preserving sampling (ds)}. When it is applied on MOEA/D-GM, to emphasize it, the algorithm is called MOEA/D-GM-ds.

\section{Multi-objective Deceptive Optimization Problem} \label{sec:DECEPTION}

There exists a class of scalable problems where the difficulty is given by the interactions that arise among subsets of decision variables. Thus, some problems should require the algorithm to be capable of linkage learning, i.e., identifying and exploring interactions between the decision variables to provide effective exploration. Decomposable deceptive problems \cite{Goldberg:1987} have played a fundamental role in the analysis of EAs. As mentioned before, one of the advantages of EDAs that use probabilistic graphical models is their capacity to capture the structure of deceptive functions. Different works in the literature have proposed EDAs \cite{Deb_and_Goldberg:1991,Sastry_et_al:2005,Echegoyen_et_al:2007,Echegoyen_et_al:2012a} and MO-EDAs \cite{Pelikan_et_al:2005,Martins_et_al:2011} for solving decomposable deceptive problems. 
 
   One example of this class of decomposable deceptive functions is the \textit{Trap-k}, where $k$ is the fixed number of variables in the subsets (also called, partitions or \textit{building blocks}) \cite{Deb_and_Goldberg:1991}. Traps deceive the algorithm away from the optimum if interactions between the variables in each partition are not considered. According to \cite{Pelikan_et_al:2005}, that is why standard crossover operators of genetic algorithms fail to solve traps unless the bits in each partition are located close to each other in the chosen representation. 
  
 Pelikan et al \cite{Pelikan_et_al:2005} used a bi-objective version of \textit{Trap-k} for analyzing the behavior of multi-objective hierarchical BOA (hBOA). 
  
  The functions $f_{trap5}$ (Equation~\eqref{eq:TRAP5}) and $f_{inv\_trap5}$ (Equation~\eqref{eq:INVTRAP5}) consist in evaluating a vector of decision variables $\mathbf{x} \in \{0,1\}^{n}$, in which the positions of $\mathbf{x}$ are divided into disjoint subsets or partitions of 5 bits each ($n$ is assumed to be a multiple of 5). The partition is fixed during the entire optimization run, but the algorithm is not given information about the partitioning in advance. Bits in each partition contribute to \textit{trap} of order 5 using the following functions:
  \begin{eqnarray}
 f_{trap5}(\mathbf{x}) = \sum^{l-1}_{k=0}\sum^{5}_{i=1}trap5(x_{5*k+i}) \label{eq:TRAP5} \\ 
 trap5(u) =
  \begin{cases}
  5          & \text{if } u = 5 \\
   4-u       & \text{if } u < 5 \nonumber \label{eq:TRAP}
  \end{cases} 
  \end{eqnarray}

  \begin{eqnarray}
 f_{inv\_trap5}(\mathbf{x}) = \sum^{l-1}_{k=0}\sum^{5}_{i=1}inv\_trap5(x_{5*k+i}) \label{eq:INVTRAP5} \\ 
 inv\_trap5(u) =
  \begin{cases}
  5          & \text{if } u = 0 \\
   u-1       & \text{if } u >0   \nonumber \label{eq:INVTRAP}
  \end{cases}
  \end{eqnarray} where $l$ is the number of building blocks, i.e., $n=5l$, and $u$ is the number of ones in the input string of 5 bits.

In the bi-objective problem, $f_{trap5}$ and $f_{inv\_trap5}$ are conflicting. Thus, there is not one single global optimum solution but a set of Pareto optimal solutions. Moreover, the amount of possible solutions grows exponentially with the problem size \cite{Sastry_et_al:2005}. Multi-objective \textit{Trap-k} has been previously investigated for Pareto based multi-objective EDAs \cite{Pelikan:2005,Pelikan_et_al:2005,Martins_et_al:2011}.  

\section{Related work} \label{sec:RELWORK}

In this section we review a number of related works emphasizing the differences to the work presented in this paper. Besides, we did not find any previous report of multi-objective decomposition approaches that incorporate PGMs for solving combinatorial MOPs.


\subsection{MOEA/D using univariate EDAs}

 In \cite{Zhou:2013}, the multi-objective estimation of distribution algorithm based on decomposition (MEDA/D) is proposed for solving the multi-objective Traveling Salesman Problem (mTSP). For each subproblem $i$, MEDA/D uses a matrix $Q^i$ to represent the  connection "strength" between cities in the best solutions found so far. Matrix $Q^i$ is combined with a priori information about the distances between cities of problem $i$, in a new matrix $P^i$ that represents  the probability that the $s$th and $t$th cities are connected in the route of the ith sub-problem. Although each matrix $P^i$ encodes a set of probabilities relevant for each corresponding TSP subproblem, these matrices can not be considered PGMs since they  do not comprise a graphical component representing the dependencies of the problem. The type of updates applied to the matrices is more related to parametric learning than to structural learning as done in PGMs.  Furthermore, this type of "models" resemble  more the class of structures traditionally used for ACO  and they  heavily depend on the use of prior information (in the case of MEDA/D, the incorporation of information about the distances between cities).  Therefore, we do not consider MEDA/D as a member of the MOEA/D-GM class of algorithms.

 Another approach combining the use of probabilistic models and MOEAD for mTSP is presented in  \cite{Shim:2012}.  In that paper, a univariate model is used to encode the probabilities of each city of the TSP being assigned to each  of the possible positions of the permutation. Therefore, the model is represented as a matrix of dimension $n \times n$ comprising the univariate probabilities of the city configurations for each position.  One main difference of this hybrid MEDA/D with our approach, and with the proposal of Zhou et al \cite{Zhou:2013}, is that a single matrix is learned using all the solutions. Therefore, the information contained in the univariate model combines information from all the subproblems, disregarding the potential regularities contained in the local neighborhoods. Furthermore, since the sampling process from the matrix does not take into account the constraints related to the permutations, repair mechanisms and penalty functions are used to "correct" the infeasible solutions. As a consequence, much of the information sampled from the model to the solution can be modified by the application of the repair mechanism.
 
These previous MEDA/Ds were applied for solving permutation-based MOPs. The application of EDAs for solving permutation problems is increasing in interest. A number of EDAs have been also specifically designed to deal with permutation-based problems \cite{Ceberio_et_al:2012, Ceberio_et_al:2013}.  The framework proposed in this paper is only investigated for binary problems. However, EDAs that deal with the permutation space can be incorporated into the framework in the future.
 
  In \cite{Wang:2015}, the authors proposed a univariate MEDA/D for solving the (binary) multi-objective knapsack problem (MOKP) that uses an adaptive operator at the sampling step to preserve diversity, i.e., prevents the learned probability vector from premature convergence. Therefore, the sampling step depends on both the univariate probabilistic vector and an extra parameter "$r$".
 

\subsection{MOEA/D using multivariate EDAs}

 In \cite{Giagkiozis_et_al:2012}, a decomposition-based algorithm is proposed to solve many-objective optimization problems. The proposed framework (MACE-gD) involves two ingredients: (i) a concept called generalized decomposition, in which the decision maker can guide the underlying search algorithm toward specific regions of interest, or the entire Pareto front and (ii) an EDA based on low-order statistics, namely the cross-entropy method \cite{Botev_et_al:2013}. MACE-gD is applied on a set of many-objective continuous functions. The obtained results showed that the proposed algorithm is competitive with the standard MOEA/D and RM-MEDA \cite{Zhang_et_al:2008}. The class of low-order statistics used by MACE-gD (Normal univariate models) limit the ability of the algorithm to capture and represent interactions between the variables. The univariate densities are updated using an updating rule as the one originally proposed for the PBIL algorithm \cite{Baluja:1994}.
 
  In \cite{Zapotecas_et_al:2015},  the covariance matrix adaptation evolution strategy (CMA-ES) \cite{Hansen_and_Ostermeier:1996} is used as the probabilistic model of MOEA/D. Although CMA-ES was introduced and has been developed in the context of evolutionary strategies, it learns a Gaussian model of the search. The covariance matrix learned by CMA-ES is able to capture dependencies between the variables. However, the nature of probabilistic modeling in the continuous domain is different to the one in the discrete domain. The methods used for learning and sampling the models are different. Furthermore, the stated main purpose of the work presented in \cite{Zapotecas_et_al:2015} was \emph{to investigate to what extent the CMA-ES could be appropriately integrated in MOEA/D and what are the benefits one could obtain}. Therefore, emphasis was put on the particular adaptations needed by CMA-ES to efficiently learn and sample its model in this different context. Since these adaptations are essentially different that the ones required by the discrete EDAs used in this paper, the contributions are different. 

\subsection{Other MO-EDAs}

Pelikan et al. \cite{Pelikan_et_al:2006a} discussed the multi-objective decomposable problems and their difficulty. The authors attempted to review a number of MO-EDAs, such as:  multi-objective mixture-based iterated density estimation algorithm (mMIDEA) \cite{Thierens_and_Bosman:2001}, multi-objective mixed Bayesian optimization algorithm (mmBOA) \cite{Laumanns_and_Ocenasek:2002} and multi-objective hierarchical BOA (mohBOA) \cite{Khan:2003}. Moreover, the authors introduced an improvement to mohBOA. The algorithm combines three ingredients: (i) the hierarchical Bayesian optimization algorithm (hBOA) \cite{Khan:2003}, (ii) the multi-objective concepts from NSGAII \cite{Deb_et_al:2002} and (iii) clustering in the objective space. The experimental study showed that the mohBOA efficiently solved multi-objective decomposable problems with a large number of competing \textit{building blocks}. The algorithm was capable of effective recombination by building and sampling Bayesian networks with decision trees, and significantly outperformed algorithms with standard variation operators on problems that require effective linkage learning.

All MO-EDAs covered in \cite{Pelikan_et_al:2006a} are Pareto-based and they use concepts from algorithms such as NSGAII and SPEA2 \cite{Laumanns_and_Ocenasek:2002}. Since, in the past few years, the MOEA/D framework has been one of the major frameworks to design MOEAs \cite{MOEADforMany:2014}, incorporating probabilistic graphical models into MOEA/D seems to be a promising technique to solve scalable deceptive multi-objective problems.

 Martins et al. \cite{Martins_et_al:2011} proposed a new approach for solving decomposable deceptive multi-objective problems. The MO-EDA, called mo$\phi$GA, uses a probabilistic model based on a phylogenetic tree. The mo$\phi$GA was tested on the multi-objective deceptive functions $f_{trap5}$ and $f_{inv\_trap5}$. The mo$\phi$GA outperformed mohBOA in terms of number of function evaluations to achieve the exact \textit{PF}, specially when the problem in question increased in size. 

 A question discussed in \cite{Martins_et_al:2011} is: if such probabilistic model can identify the correct correlation between the variables of a problem, the combination of improbable values of variables can be avoided. However, as the model becomes more expressive, the computational cost of incurred by the algorithm to build the model also grows. Thus, there is a trade-off between the efficiency of the algorithm for building models and its accuracy. In our proposal, at each generation, $N$ probabilistic graphical models are kept. Therefore, a higher number of subproblems $N$ can be a drawback for the proposal in terms of efficiency (time consuming). However, if the proposed MOEA/D-GM has an adequate commitment between the efficiency and accuracy (approximated \textit{PF}) then it is expected to behave satisfactorily.

\subsection{Contributions with respect to previous work}

We summarize some of the main contributions of our work with respect to the related research.
\begin{itemize}
  \item We use, for the first time, a probabilistic graphical model within MOEA/D to solve combinatorial MOPs. In this case, the previous MOEA/Ds that incorporate probabilistic models  cover only univariate models.
  \item We investigate a particular class of problems (deceptive MOPs) for which there exist extensive evidence about the convenience of using probabilistic graphical models. 
  \item We introduced in MOEA/D the class of probabilistic models where the structure of the model is learned from the neighborhood of each solution. 
  \item We investigate the question of how the problem interactions are kept in the scalarized subproblems and how these interactions are translated to the probabilistic models. 
\end{itemize}

\section{Experiments} \label{sec:EXPE}

The proposed MOEA/D-GM framework has been implemented in C++. In the comparison study, four different instantiations are used. For convenience, we called each algorithm instantiated as: MOEA/D-GA, MOEA/D-UMDA,  MOEA/D-PBIL and MOEA/D-Tree. 

To evaluate the algorithms, the deceptive functions $f_{trap5}$ and $f_{inv\_trap5}$ (Section \ref{sec:DECEPTION}) are used to compose the bi-objective \textit{Trap} problem with different number of variables $n \in \{30, 50, 100\}$. Equation~\eqref{eq:bitrap} defines the notation of the \textit{bi-Trap} function.

\begin{eqnarray}
   bi\textit{-}Trap(\mathbf{x}) = (f_{trap5}(\mathbf{x}), f_{inv\_trap5}(\mathbf{x})) \label{eq:bitrap} \\ 
subject \ to \ \mathbf{x} \in \{0,1\}^{n}  \nonumber
\end{eqnarray}

\subsection{Performance metrics}

The \textit{true PF} from \textit{bi-Trap} is known. Therefore, two performance metrics are used to evaluate the algorithms performance: (i) inverted generational distance metric (IGD) \cite{Zitzler_and_Thiele:1999}, and (ii) the number of \textit{true} Pareto optimal solutions found by the algorithm through the generations. The stop condition for all the algorithms is $5n$ generations. 


 
 Let $P^{*}$ be a set of uniformly distributed Pareto optimal solutions along the \textit{true} $PF$ in the objectives. $P$ be an \textbf{approximated} set to the \textit{true} $PF$ obtained by an algorithm. 
 
\textbf{IGD-metric} \cite{Zitzler_and_Thiele:1999}: The inverted generational distance from $P^{*}$ to $P$ is defined as 
\begin{equation}
IGD(P^{*},P) = \frac{\sum^{v \in P^{*}}d(v,P)}{|P^{*}|}
\end{equation}
where $d(v,P)$ is the minimum Euclidean distance between $v$ and the solutions in $P$. If $|P^{*}|$ is large enough $IGD(P^{*},P)$ could measure both convergence and diversity of $P$ in a sense. The lower the $IGD(P^{*},P)$, the better the approximation of $P^{*}$ to the true \textit{PF}. 

\textbf{Number of \textit{true} Pareto optimal solutions:} The number of Pareto solutions that composes $P$ that belongs to the $P^{*}$ is defined as
\begin{equation}
	|P^{+}| = | P^{*} \cap P |
\end{equation}
where $|P^{+}|$ is the cardinality of $P^{+}$.

Also, we use the statistical test \textit{Kruskall-Wallis} \cite{Kruskal} to rank the algorithms according to the results obtained by the metrics. If two or more algorithms achieve the same rank, it means that there is no significant difference between them. 

\subsection{Parameters Settings}

\begin{table*}[]
\centering \scriptsize
\caption{ \small Results from the average IGD and average number of \textit{true} Pareto optimal solutions $|P^{+}|$ computed form the 30 runs for each combination (problem size $n \times$ Algorithm $\times$ preserving sampling). The column $|P^{*}|$ is the total number of \textit{true} Pareto solutions for each problem $n$.}
\label{average}
\begin{tabular}{@{\extracolsep{\fill} }llllll|llll}
\hline \noalign{\smallskip}
\multicolumn{10}{c}{Average number of true Pareto optimal solutions $|P^{+}|$}                                                                                                                                                                                                                                                                                                               \\ \hline
\multicolumn{1}{c}{\multirow{2}{*}{$n$}} & \multicolumn{1}{c}{\multirow{2}{*}{$|P^{*}|$}} & \multicolumn{4}{c|}{standard sampling}                                                                                               & \multicolumn{4}{c}{diversity preserving sampling (ds)}                                                                                                       \\
\multicolumn{1}{c}{}                              & \multicolumn{1}{c}{}                           & \multicolumn{1}{c}{MOEA/D-GA} & \multicolumn{1}{c}{MOEA/D-PBIL} & \multicolumn{1}{c}{MOEA/D-UMDA} & \multicolumn{1}{c|}{MOEA/D-Tree} & \multicolumn{1}{c}{MOEA/D-GA} & \multicolumn{1}{c}{MOEA/D-PBIL} & \multicolumn{1}{c}{MOEA/D-UMDA} & \multicolumn{1}{c}{MOEA/D-Tree}    \\ \hline \noalign{\smallskip}
30                                                & 7                                              & \multicolumn{1}{l}{4.167}     & \multicolumn{1}{l}{3.034}       & \multicolumn{1}{l}{2.767}       & \multicolumn{1}{l|}{6.067}       & \multicolumn{1}{l}{6.267}     & \multicolumn{1}{l}{4.034}       & \multicolumn{1}{l}{4.134}       & \multicolumn{1}{l}{\textbf{6.9}}   \\
50                                                & 11                                             & \multicolumn{1}{l}{3.867}     & \multicolumn{1}{l}{2.567}       & \multicolumn{1}{l}{3.034}       & \multicolumn{1}{l|}{5.134}       & \multicolumn{1}{l}{6.634}     & \multicolumn{1}{l}{2.867}       & \multicolumn{1}{l}{3.2}         & \multicolumn{1}{l}{\textbf{9.5}}   \\
100                                               & 21                                             & \multicolumn{1}{l}{3.3}       & \multicolumn{1}{l}{2.534}       & \multicolumn{1}{l}{3.5}         & \multicolumn{1}{l|}{2.967}       & \multicolumn{1}{l}{5.334}     & \multicolumn{1}{l}{2.8}         & \multicolumn{1}{l}{3.433}       & \multicolumn{1}{l}{\textbf{9.367}} \\ \hline \noalign{\smallskip}
\multicolumn{10}{c}{average IGD measure $IGD(P^{*},P)$}                                                                                                                                                                                                                                                                                                                                           \\ \hline \noalign{\smallskip}
30                                                & 7                                              & 1.076                         & 1.764                           & 1.853                           & 0.418                            & 0.371                         & 1.258                           & 1.246                           & \textbf{0.053 }                             \\
50                                                & 11                                             & 2.174                         & 3.585                           & 3.229                           & 1.657                            & 1.308                         & 3.4                             & 3.158                           & \textbf{0.501 }                             \\
100                                               & 21                                             & 4.757                         & 7.663                           & 6.38                            & 4.815                            & 4.142                         & 7.473                           & 6.726                           &        \textbf{3.359 }                           \\ \hline \noalign{\smallskip}
\end{tabular}
\end{table*}

\begin{table*}[]
\centering \scriptsize
\caption{\small Kruskall Wallis statistical ranking test according to the IGD and $|P^{+}|$ for each combination (problem size $n \times$ Algorithm $\times$ preserving sampling). The first value is the average rank over the 240 runs (i.e., 30 runs $\times$ 8 algorithms). The second value (in brackets) is the final ranking from 8 to 1. If two or more ranks are the same, it means that there is no significant difference between them.}
\label{kruskal}
\begin{tabular}{@{\extracolsep{\fill} }llrrrr|rrrr} 
\hline \noalign{\smallskip}
\multicolumn{10}{c}{Kruskall-Wallis ranking - number of true Pareto optimal solutions $|P^{+}|$}                                                                                                                                                                                                                                                                             \\ \hline \noalign{\smallskip}
\multicolumn{1}{c}{\multirow{2}{*}{n}} & \multicolumn{1}{c}{\multirow{2}{*}{}} & \multicolumn{4}{c|}{standard sampling}                                                                                               & \multicolumn{4}{c}{diversity preserving sampling (ds)}                                                                                                       \\
\multicolumn{1}{c}{}                   & \multicolumn{1}{c}{}                      & \multicolumn{1}{c}{MOEA/D-GA} & \multicolumn{1}{c}{MOEA/D-PBIL} & \multicolumn{1}{c}{MOEA/D-UMDA} & \multicolumn{1}{c|}{MOEA/D-Tree} & \multicolumn{1}{c}{MOEA/D-GA} & \multicolumn{1}{c}{MOEA/D-PBIL} & \multicolumn{1}{c}{MOEA/D-UMDA} & \multicolumn{1}{c}{MOEA/D-Tree}    \\ \hline \noalign{\smallskip}
30                                                &                                               & 139.37 (5.50)                 & 184.88 (6.00)                   & 195.72 (6.50)                   & \textbf{66.25 (2.00)}                     & \textbf{59.43 (2.00)}         & 146.68 (6.00)                   & 141.87 (6.00)                   & \textbf{29.80 (2.00)}           \\
50                                                &                                              & 126.48 (5.00)                 & 188.30 (6.50)                   & 165.55 (6.00)                   & 88.15 (3.00)                     & 52.65 (2.00)                  & 171.40 (6.00)                   & 153.43 (6.00)                   & \textbf{18.03 (1.50)}           \\
100                                               &                                              & 136.43 (5.50)                 & 178.67 (6.00)                   & 120.37 (5.00)                   & 151.77 (5.50)                    & 61.58 (2.00)                  & 162.40 (5.50)                   & 128.37 (5.50)                   & \textbf{24.42 (1.50)}           \\ \hline \noalign{\smallskip}
\multicolumn{10}{c}{Kruskall-Wallis ranking - IGD measure}                                                                                                                                                                                                                                                                                                          \\ \hline \noalign{\smallskip}
30                                     & & 134.13 (5.50)                 & 187.05 (6.00)                   & 195.22 (6.50)                   & \textbf{65.32 (2.00)}            & \textbf{59.43 (2.00)}         & 148.67 (6.00)                   & 144.38 (6.00)                   & \textbf{29.80 (2.00)}              \\
50                                     & & 106.88 (3.50)                 & 194.43 (6.50)                   & 169.37 (6.50)                   & 74.47 (2.50)                     & 58.33 (2.50)                  & 179.35 (6.50)                   & 160.17 (6.00)                   & \textbf{21.00 (2.00)}                       \\
100                                    & & \textbf{80.30 (2.50)}         & 204.03 (7.00)                   & 145.72 (6.00)                   & \textbf{82.08 (2.50)}            & \textbf{55.45 (2.50)}         & 195.37 (6.50)                   & 162.12 (6.50)                   & \textbf{38.93 (2.50)}              \\ \hline \noalign{\smallskip}
\end{tabular}
\end{table*}

  \begin{figure*}[!ht]
\centering
\subfloat[MOEA/D-Tree]{
\includegraphics[width=8cm]{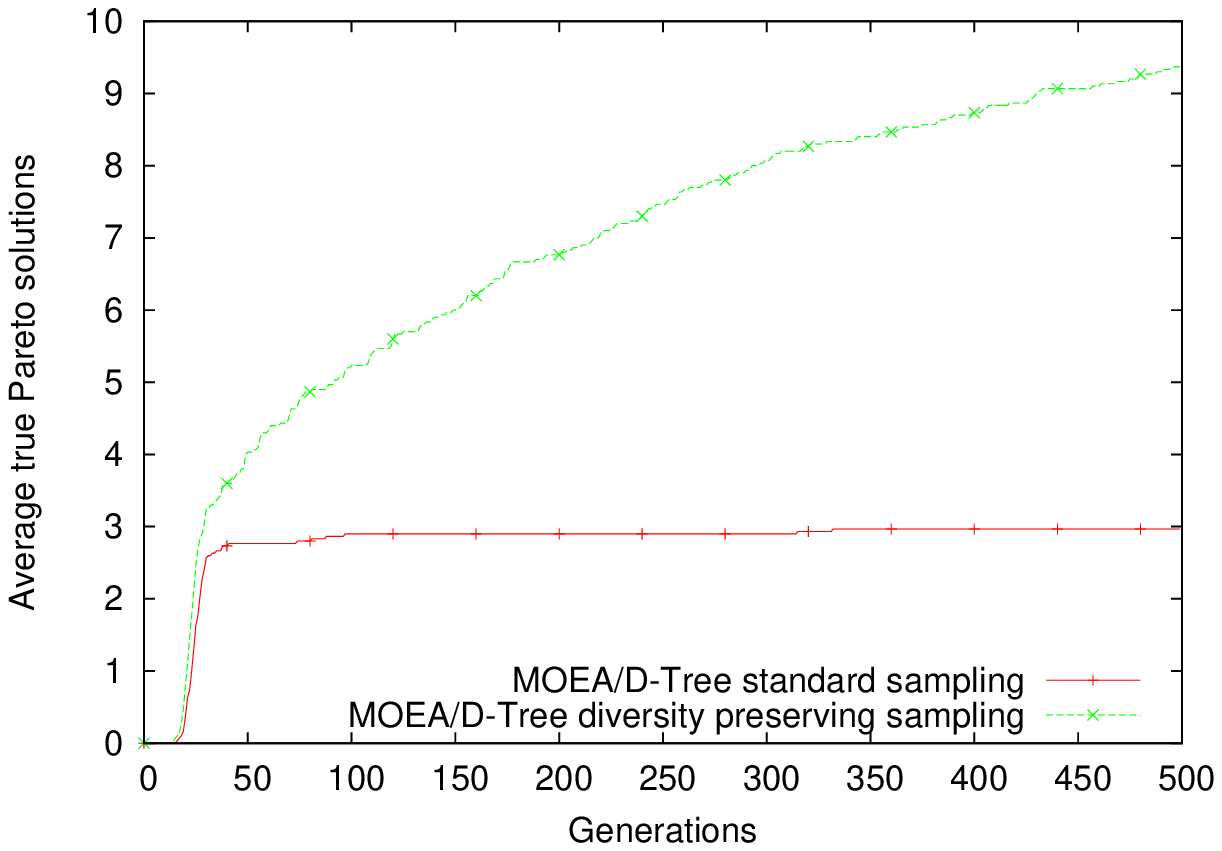}

}
\quad
\subfloat[MOEA/D-GA]{
\includegraphics[width=8cm]{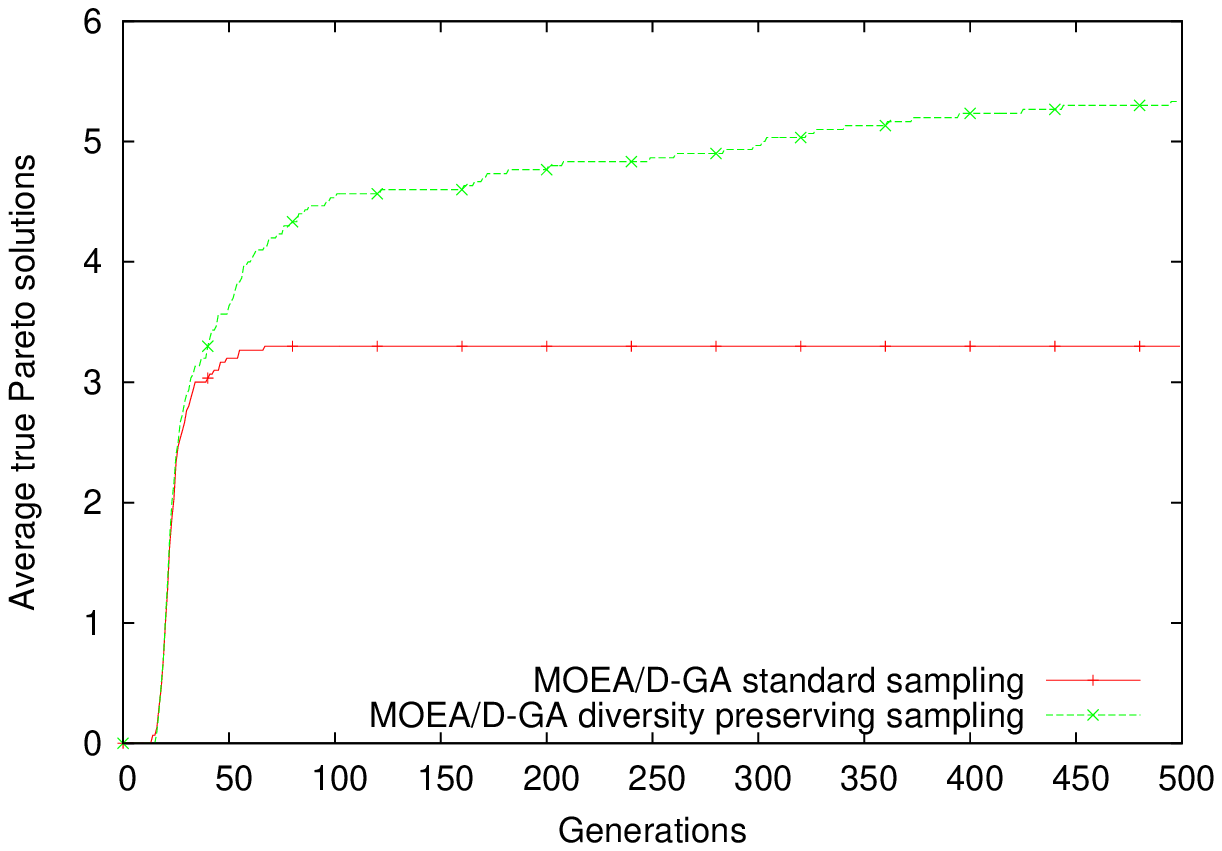}
}

\caption{Average \textit{true} Pareto optimal solutions $|P^{+}|$ over the generations. Each chart shows a comparison between the standard sampling and the diversity preserving sampling using the same algorithm.}
\label{fig1}
\end{figure*} 

The following  parameters settings were used in the experimental study:
 
\begin{enumerate}
\item \textit{Number of subproblems $(N)$:} As in most of MOEA/D algorithms proposed in the literature \cite{Zhang_and_Li:2007}, \cite{ComplicatedPareto:2008}, \cite{MOEADforMany:2014}, the number of subproblems $N$ and their correspondent weight vectors $\lambda^{1},\dots,\lambda^{N}$ are controlled by a parameter $H$, which makes a wide spread distribution of the weight vectors according to $N = C^{m-1}_{H+m-1}$. Thus, for the bi-objective problem, we set $H=200$, consequently $N=201$.

\item \textit{Neighborhood size $(T_{s})$}: As the number of selected solutions is crucial for EDAs, we test a range of neighborhood size values $T_{s}=(1, 5, 10, 20, 30, 40, 50, 60, 70, 80, 90, 100, 150)$. Also, as in \cite{Raplacement:2014}, $T_{s}=T_{r}$.

\item \textit{Maximal number of replacements by a new solution $(n_{r})$:} As in \cite{ComplicatedPareto:2008}, $n_{r}=2$.

\item \textit{Scalarization function:} We have applied both \textit{Weighted Sum} and \textit{Tchebycheff} approaches. As both achieve very similar results, only the results with \textit{Tchebycheff} are presented in this paper.

\item \textit{Maximum number of tries in diversity preserving sampling procedure:} We set this parameter with the same value of the neighborhood size $T_{s}$. 

\item \textit{Genetic operators from MOEA/D-GA:} \textit{uniform crossover} and \textit{mutation rate}$=\frac{1}{n}$.
\item \textit{PBIL learning rate:} $\alpha = 0.05$.
\end{enumerate}

Each combination of (algorithm $\times$ parameters setting) is independently run 30 times for each problem size $n$. 

 \subsection{Comparison of the different variants of MOEA/D-GM for the \textit{bi-Trap5}}
 
 \begin{figure*}[!ht]
\centering
\quad
\subfloat[$n=50$]{
\includegraphics[width=8cm]{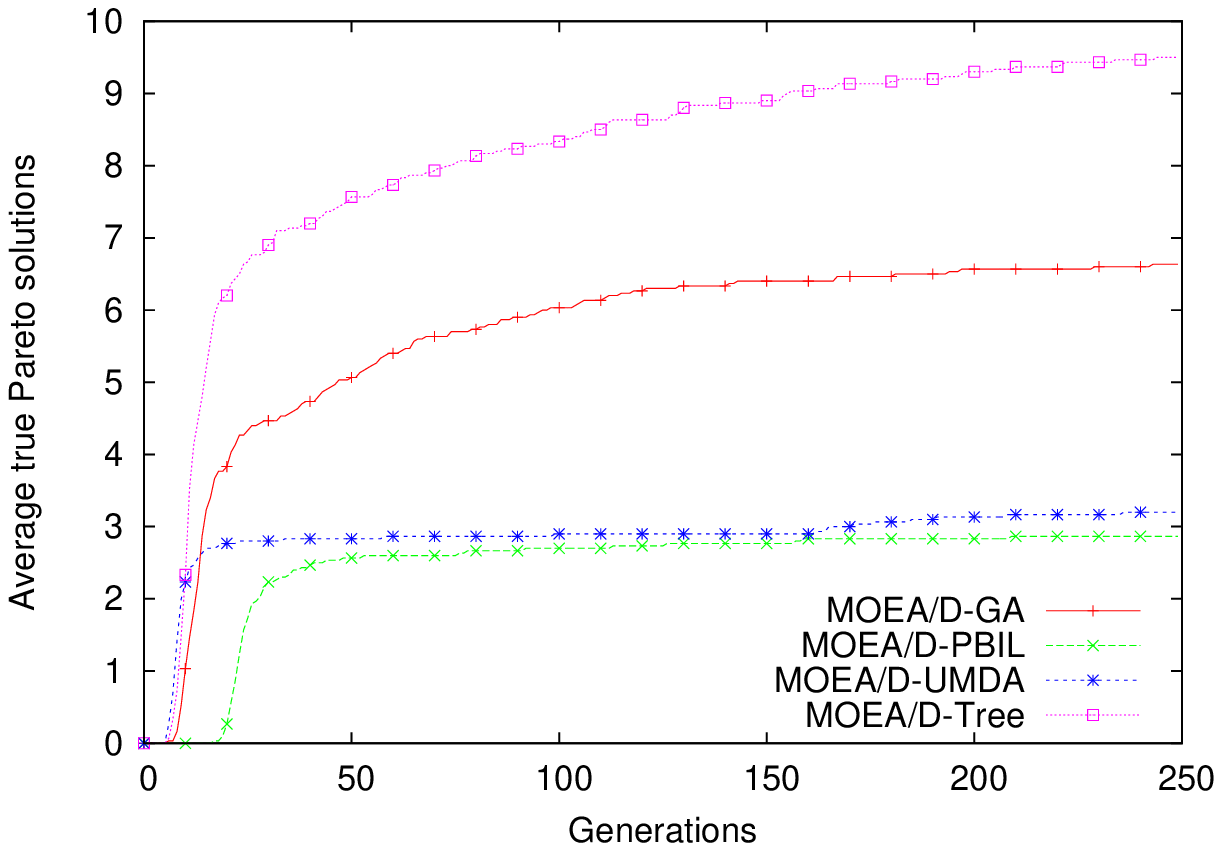}
}
\quad
\subfloat[$n=100$]{
\includegraphics[width=8cm]{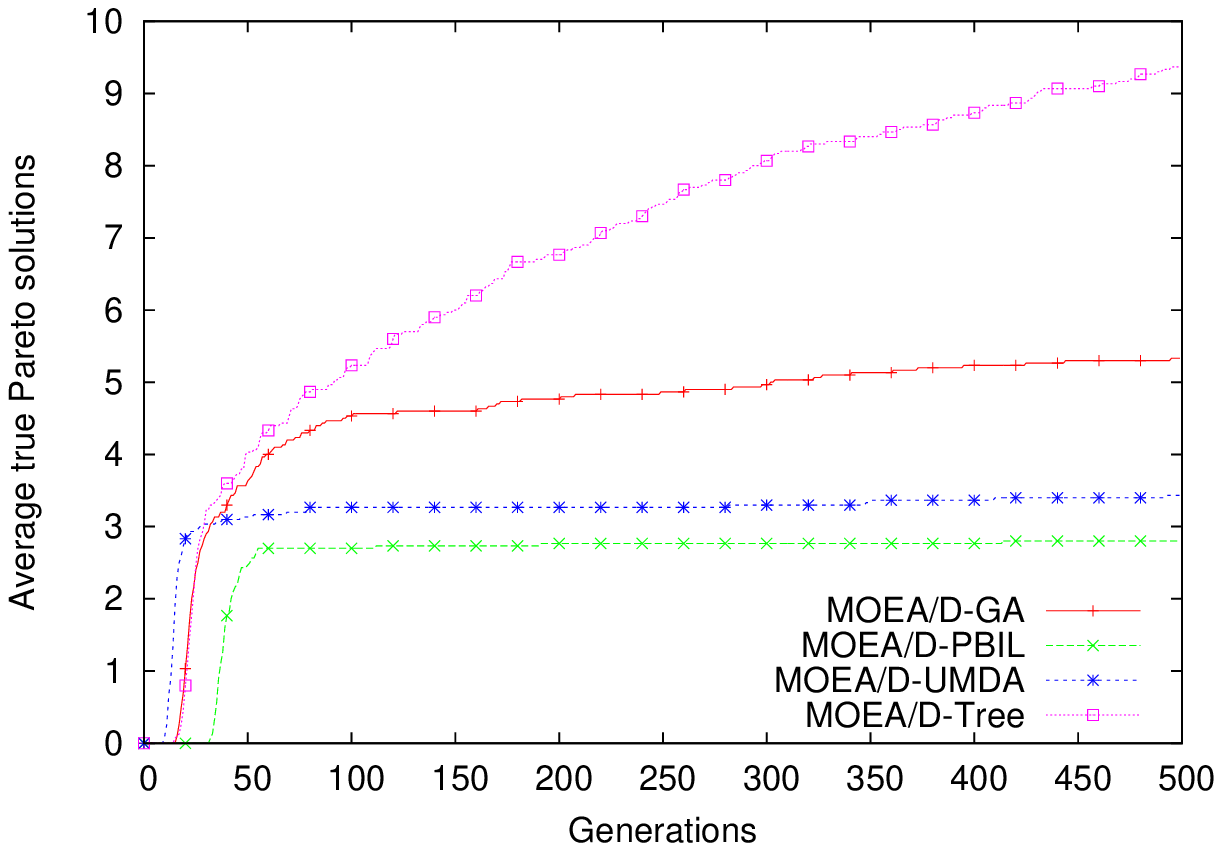}
}
\caption{Average \textit{true} Pareto solutions $|P^{+}|$ through the generations for problem size 50 and 100 with $T_{s}=20$}
\label{fig2}
\end{figure*} 

 \begin{figure*}[!ht]
\centering
\subfloat[$n=50$]{
\includegraphics[width=8cm]{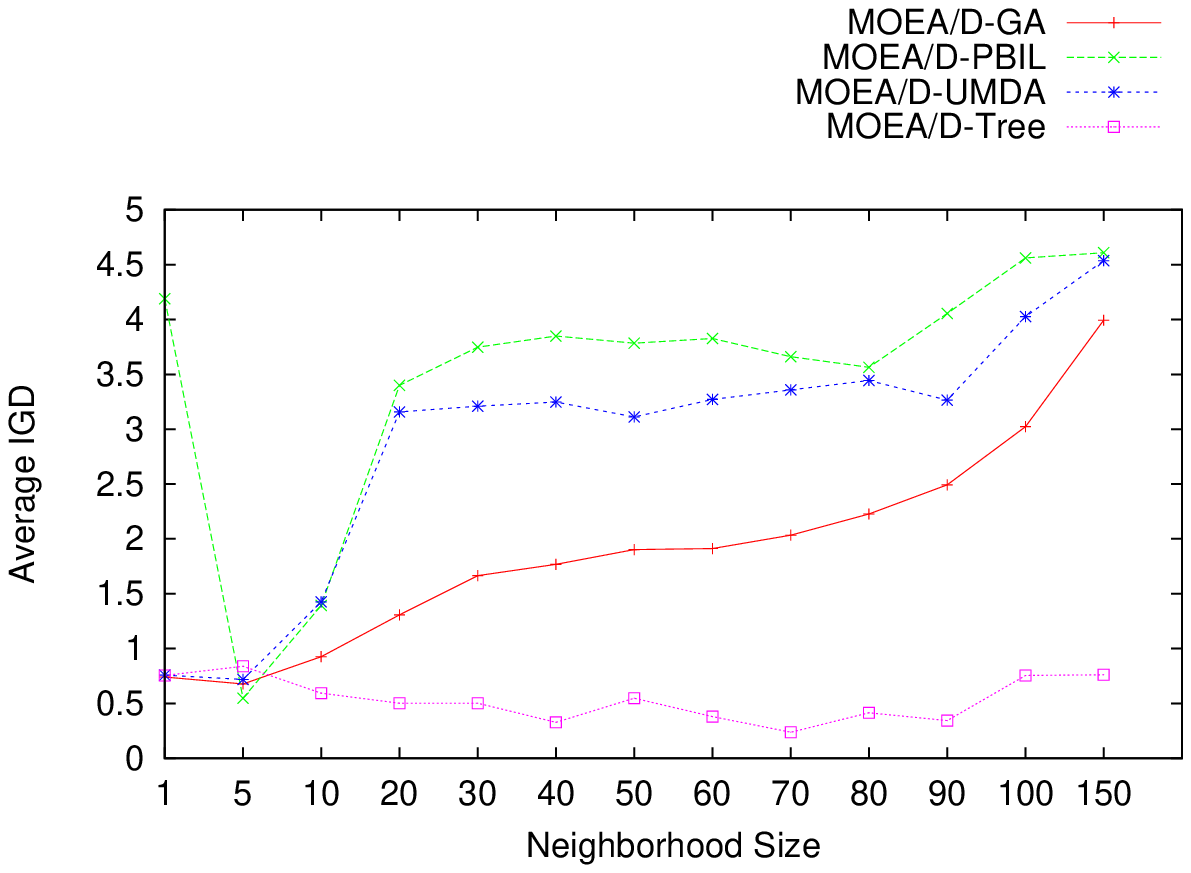}
}
\quad 
\subfloat[$n=100$]{
\includegraphics[width=8cm]{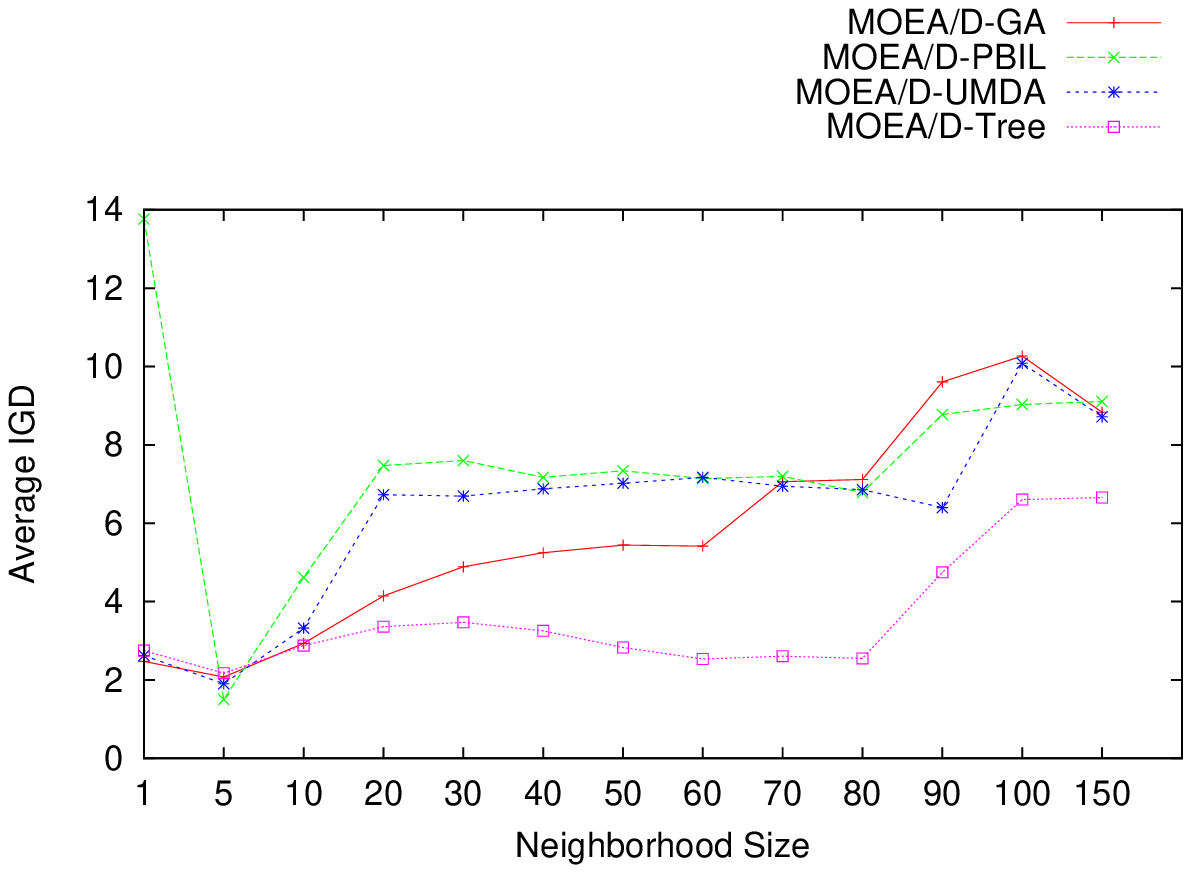}
}
\caption{Average IGD with different neighborhood sizes for selection $T_{s} \in \{1, 5, 10, 20, 30, 40, 50, 60, 70, 80, 90, 100,150\}$}
\label{fig:neibor}
\end{figure*}

 First, the algorithms are evaluated using a fixed neighborhood size value ($T_{s}= 20$). Table~\ref{average} presents the average values of the IGD and the number of \textit{true} Pareto solutions $|P^{+}|$ computed using the approximated \textit{PF} found by the algorithms. Table \ref{kruskal} presents the  results of the  \textit{Kruskall-Wallis} test (at $5\%$ significance level) applied to the results obtained by the algorithms (the same results summarized in Table~\ref{average}). The best rank algorithm(s) is(are) shown in bold. Figure~\ref{fig1} and \ref{fig2} show the behavior of the algorithms throughout the generations according to the average $|P^{+}|$ obtained.

From the analysis of the results, we can extract the following conclusions:

According to Table~\ref{kruskal}, the MOEA/D-Tree that uses the diversity preserving mechanism (MOEA/D-Tree-ds) is the only algorithm that has achieved the best rank in all the cases according to both indicators. Moreover, the diversity preserving sampling has improved the behavior of all the algorithms (mainly the MOEA/D-GA and MOEA/D-Tree) in terms of the quality of the approximation to the \textit{true PF}. Figure \ref{fig1} confirms these results. Therefore, the diversity preserving sampling has a positive effect in the algorithms for the \textit{bi-Trap5} problem, i.e., the algorithms are able to achieve a more diverse set of solutions. Additionally, Figure~\ref{fig2} shows that MOEA/D-Tree-ds achieves better results than the other algorithms from the first generations.

Moreover, all the algorithms can find, at least, one global optimal solution, i.e., a true Pareto solution. One possible explanation for this behavior is a potential advantageous  effect introduced by the clustering of the solutions determined by the MOEA/D framework. This benefit would be independent of the type of models used to represent the solutions. Grouping similar solutions in a neighborhood may allow the univariate algorithm to produce some global optima, even if the functions are deceptive. 

\subsection{Influence of the neighborhood size selection $T_{s}$ for the learning}

\begin{figure*}[!th]
\centering 
\subfloat[$T_{s}= 20$, subproblem $i=1$ ]{
\includegraphics[width=7.5cm]{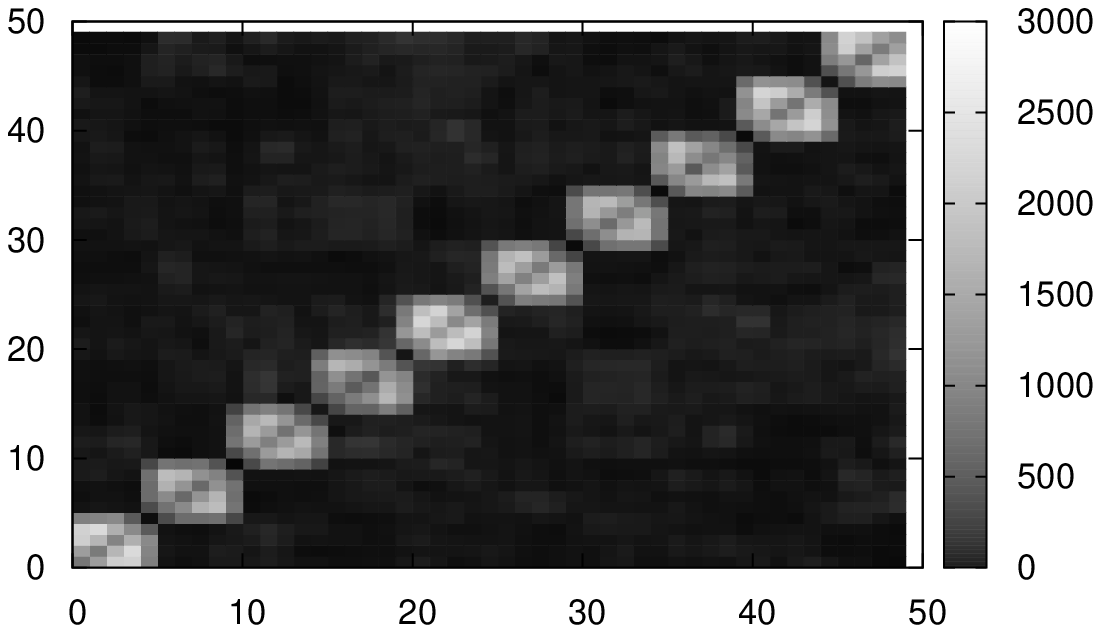}
}\vspace{-0.3cm}
\quad 
\subfloat[ $T_{s}= 70$, subproblem $i=1$]{
\includegraphics[width=7.5cm]{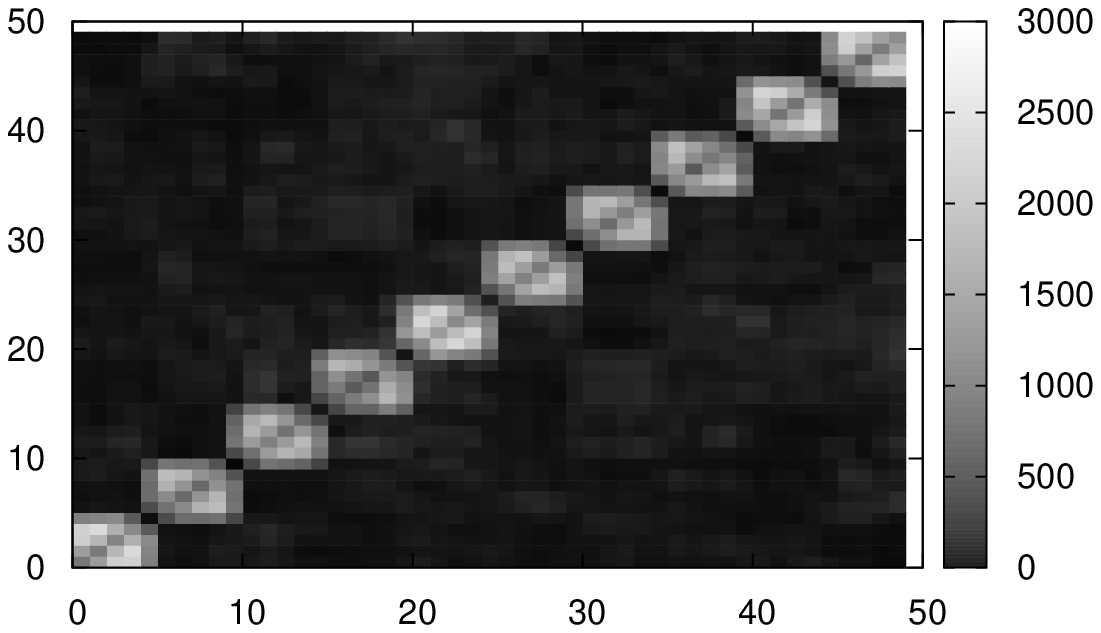}

}\vspace{-0.1cm}
\quad 
\subfloat[$T_{s}= 20$, subproblem $i=201$]{
\includegraphics[width=7.5cm]{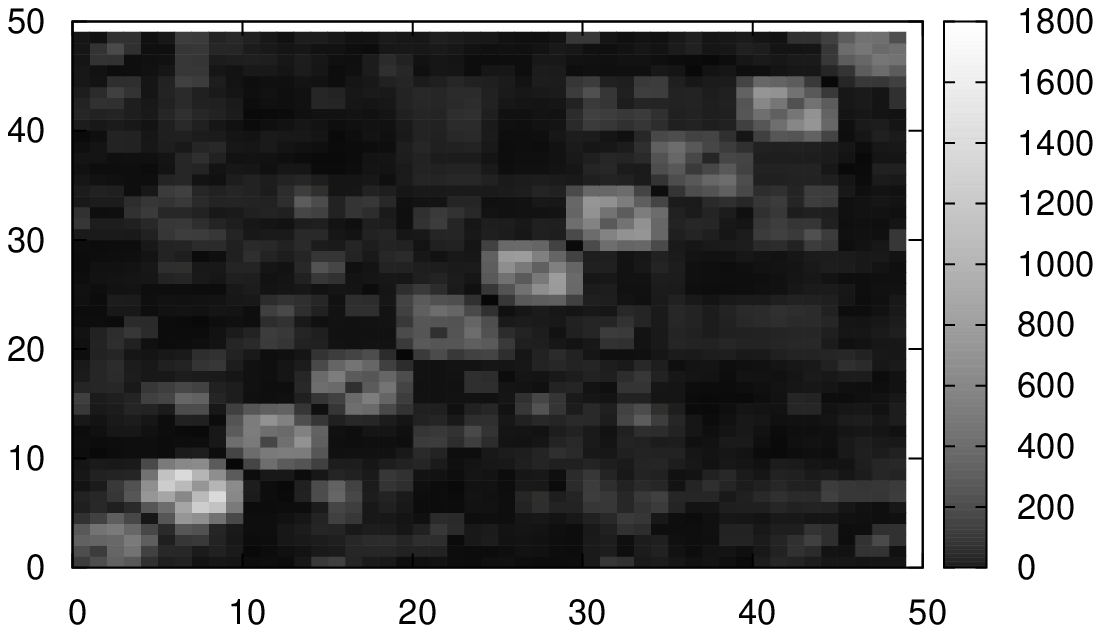}
}\vspace{-0.1cm}
\quad 
\subfloat[$T_{s}= 70$, subproblem $i=201$ ]{
\includegraphics[width=7.5cm]{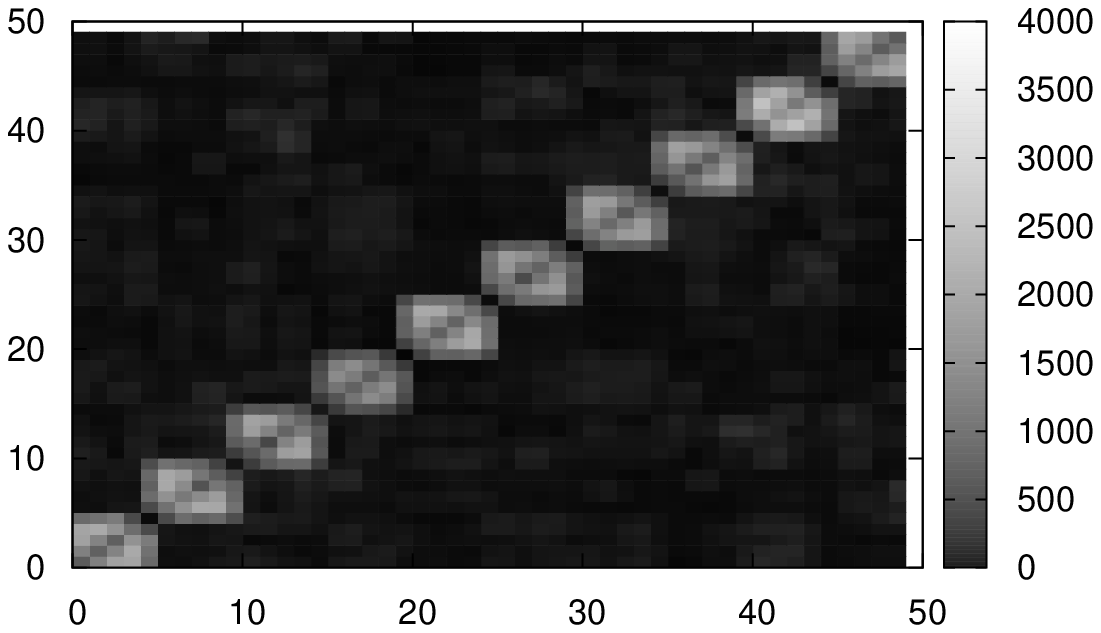}
}\vspace{-0.1cm}
\quad 
\subfloat[$T_{s}= 20$, subproblem $i=100$]{
\includegraphics[width=7.5cm]{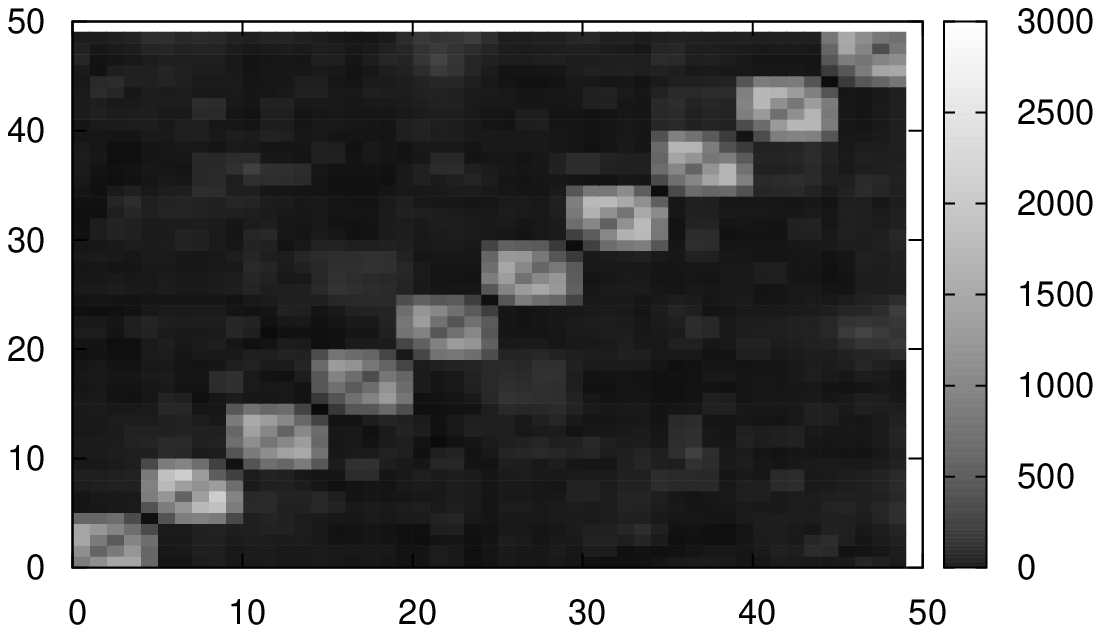}
}
\quad 
\subfloat[$T_{s}= 70$, subproblem $i=100$ ]{
\includegraphics[width=7.5cm]{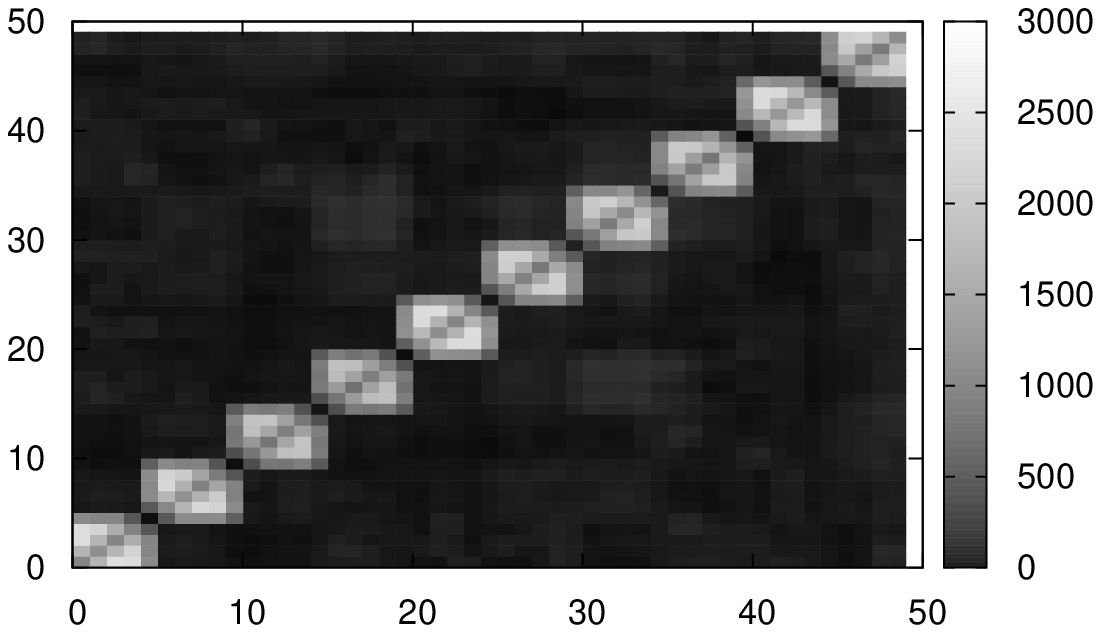}
}
\caption{Frequency matrices (\textit{heat maps}) learned by the subproblems $i=(1,100,201)$ from the MOEA/D-Tree-ds for problem size $n=50$. }
\label{frequency_matrix}

\end{figure*}
The neighborhood size has a direct impact in the search ability of the MOEA/Ds, which can balance convergence and diversity for the target problem. Figure~\ref{fig:neibor} presents the IGD values obtained with different neighborhood sizes $(T_{s})$.

As we mentioned before, the number of selected solutions has an impact for most of the EDAs. A multi-variate EDA needs a large set of selected solutions to be able to learn dependencies between the decision variables \cite{Muhlenbein_et_al:1999}. This fact may explain (see Figure~\ref{fig:neibor}) that the differences between the algorithms are lower as the neighborhood size becomes smaller (e.g., $T_{s}=5$). A remarkable point in these results is that, a small $T_{s}$ is better to solve \textit{bi-Trap5} for the algorithms except for MOEA/D-Tree-ds. MOEA/D-Tree-ds achieves good results with a large neighborhood size (e.g., 60, 70 and 80). Even if the neighborhood size has an important influence in the behavior of the algorithm, this parameter can not be seen in isolation of other parameters that also influence the behavior of the algorithm.
 
 \subsection{Analyzing the structure of problems as captured by the tree model}

One of the main benefits of EDAs is their capacity to reveal a priori unknown information about the problem structure. Although this question has been extensively studied \cite{Brownlee_et_al:2012a,Echegoyen_et_al:2007,Santana_et_al:2008a}  in the single-objective domain, the analysis in the multi-objective domain are still few \cite{Fritsche_et_al:2015,Karshenas_et_al:2014,Santana_et_al:2009}. Therefore, a relevant question was to determine to what extent is the structure of the problem captured by the probabilistic models used in MOEA/D-GM. This is a relevant question since there is no clue about the types of interactions that could be captured from  models learned for scalarized functions. In this section, the structures learned from MOEA/D-Tree-ds while solving different subproblems are investigated. 

In each generation, for each subproblem, a tree model is built according to the bi-variate probabilities obtained from its selected population. We can represent the tree model as a matrix $M_{n \times n}$, where each position $M_{jk}$ represents a relationship (\textit{pairwise}) between two variables $j,k$. $M_{jk}=1$ if $j$ is the parent of $k$ in tree model learned, otherwise $M_{jk}=0$. 

Figure~\ref{frequency_matrix} represents the merge frequency matrices obtained by the 30 runs. The frequencies are represented using \textit{heat maps}, where lighter colors indicate a higher frequency. We have plotted the merge matrices learned from the extreme subproblems and the middle subproblem i.e., $i=(1, 100, 201)$ using the problem size $n=50$.  Because of the results of the previous section, the frequency matrices for two neighborhood sizes are presented, $T_{s}=(20,70)$.

 The matrices clearly show a strong relationship between the subsets of variables "\textit{building blocks}" of size 5, which shows that the algorithm is able to learn the structure of the \textit{Trap5} functions. Notice that, for neighborhood size $70$, the MOEA/D-Tree-ds was able to capture more accurate structures, which exalts the good results found in accordance with the IGD metric. This can be explained by the fact that a higher population size reduces the number of spurious correlations learned from the data. 
 
 Moreover, analyzing the different scalar subproblems, we can see that, the algorithm is able to learn a structure even for the middle scalar subproblem $i=100$, where the two conflict objectives functions $f_{trap5}$ and $f_{inv\_trap5}$ compete in every partition (\textit{building block}) of the decomposable problem. 
 
\section{Conclusions and future work} \label{sec:CONCLU}

In this paper, a novel and general MOEA/D framework able to instantiate probabilistic graphical models named MOEA/D-GM has been introduced. PGMs are used to obtain a more comprehensible representation of a search space. Consequently, the algorithms that incorporate PGMs can provide a model expressing the regularities of the problem structure as well as the final solutions. The PGM investigated in this paper takes into account the interactions between the variables by learning a maximum weight spanning tree from the bi-variate probabilities distributions. An experimental study on a bi-objective version of a well known deceptive function (\textit{bi-Trap5}) was conducted. In terms of accuracy (approximating of the true \textit{PF}), the results have shown that the instantiation from MOEA/D-GM, called MOEA/D-Tree, is significantly better than MOEA/D that uses univariate EDAs and traditional genetic operators. 

Moreover, other enhancement has been introduced in MOEA/D. A new simple but effective mechanism of sampling has been proposed, called diversity preserving sampling. Since sampling solutions already present in the neighbor solutions can have a detrimental effect in terms of convergence, the diversity preserving sampling procedure tries generating those solutions that are different from the parent solutions. According to both performance indicators, all the algorithms in the comparison have improved their results in terms of diversity to the approximation \textit{PF}.

 An analysis of the influence of the neighborhood size on the behavior of the algorithms were conducted. In general, increasing the neighborhood size has a detrimental effect. Although, this is not always the case for MOEA/D-Tree. 
 
 Also, independent of the type of models used to represent the solutions, grouping similar solutions in a neighborhood may allow to produce some global optima, even if the functions are deceptive.

Finally, we have also investigated for the first time to what extent the models learned by MOEA/D-GMs can capture the structure of the objective functions. The analysis has been conducted for MOEA/D-Tree-ds considering the structures learned with different neighborhood sizes and different subproblems. We have found that even if relatively small neighborhoods are used (in comparison with the standard population sizes used in EDAs), the models are able to capture the interactions of the functions. One potential application of this finding, is that, we could reuse or transfer models between subproblems, in a similar way to the application of structural transfer between related problems \cite{Santana_et_al:2012f}. 

The scalability of MOEA/D-Tree-ds for the \textit{bi-Trap5} problem and other multi-objective deceptive functions should be investigated. The "most appropriate" model for \textit{bi-Trap5} should be able to learn higher order interactions (of order $4$). Therefore, PGMs based on Bayesian or Markov networks could be of application in this case. 

The directions for future work are: (i) Conceive strategies to avoid learning a model for each subproblem. This would improve the results in terms of computational cost; (ii) Use of the most probable configurations of the model to speed up convergence; (iii) Consider the application of hybrid schemes incorporating local search that take advantage of the information learned by the models; (iv) Evaluate MOEA/D-GM on other benchmarks of deceptive MOPs.

\section*{Acknowledgments}

This work has received support by CNPq (Productivity Grant Nos. 305986/2012-0 and Program Science Without Borders Nos.: 400125/2014-5), by CAPES (Brazil Government), by IT-609-13 program (Basque Government) and TIN2013-41272P (Spanish Ministry of Science and Innovation).
\bibliographystyle{IEEEtran}



\end{document}